\documentclass[review]{elsarticle}

\usepackage{lineno,hyperref}
\usepackage{amsmath}
 
\newtheorem{myTheo}{Lemma}

\usepackage{mathrsfs}
\usepackage{algorithmic}
\usepackage{bm}
\usepackage{algorithm}
\usepackage{booktabs}
\usepackage[caption=false]{subfig}
\usepackage{setspace}
\usepackage{amsfonts,amssymb}
\usepackage{graphicx}
\usepackage{color}
\usepackage{multirow}

\modulolinenumbers[5]

\journal{Journal of \LaTeX\ Templates}

\begin{document}

%
%
%

\newpage

\begin{frontmatter}

\title{Learning Linear Dynamical Systems with High-Order Tensor Data for Skeleton based Action Recognition}



\author[add1]{Wenwen Ding}
\ead{dww2048@163.com}

\author[add1]{Kai Liu}
\ead{kailiu@mail.xidian.edu.cn}

\address[add1]{School of Computer Science and Technology, Xidian University, Xi'an, China}

\begin{abstract}
In recent years, there has been renewed interest in developing methods for skeleton-based human action recognition. A skeleton sequence can be naturally represented as a high-order tensor time series. In this paper, we model and analyze tensor time series with Linear Dynamical System (LDS) which is the most common for encoding spatio-temporal time-series data in various disciplines dut to its relative simplicity and efficiency. However, the traditional LDS treats the latent and observation state at each frame of video as a column vector. Such a vector representation fails to take into account the curse of dimensionality as well as valuable structural information with human action. Considering this fact, we propose generalized Linear Dynamical System (gLDS) for modeling tensor observation in the time series and employ Tucker decomposition to estimate the LDS parameters as action descriptors. Therefore, an action can be represented as a subspace corresponding to a point on a Grassmann manifold. Then we perform classification using dictionary learning and sparse coding over Grassmann manifold. Experiments on MSR Action3D Dataset, UCF Kinect Dataset and Northwestern-UCLA Multiview Action3D Dataset demonstrate that our proposed method achieves superior performance to the state-of-the-art algorithms.
\end{abstract}


\end{frontmatter}

\section{Introduction}

Human action recognition and behavior analysis based on spatio-temporal data are one of the hottest topics in the field of computer vision due to its many applications in smart surveillance, web-video search and retrieval, human-computer interfaces, and health-care. After the recent release of cost-effective depth sensors, we witness another growth of research on 3D data. The introduction of 3D data largely alleviates the low-level difficulties of illumination changes, occlusions, background extraction,   Furthermore, the 3D positions of skeletal joints will be quickly and accurately estimated from a single depth image \cite{shotton2013real}. These recent advances have resulted in a keen interest in skeleton-based human action recognition. 

The coupling of the spatial texture and the temporal dynamics is more challenging for understanding the human action than static data. The global dynamic of action sequences is usually captured by modeling the temporal variations with Linear Dynamical System (LDS) \cite{turaga2010statistical}. The traditional model treats the latent and observation state at each human skeleton as a vector. Such vector representation fails to match the structural properties of the skeleton. In the most of  previous works \cite{aggarwal2014human,presti20163d}, representation for skeleton-based action recognition usually concatenates all the attribute of skeletal joint points together to get a single vector. In contrast, we consider a skeleton as directed graph, which the nodes are joint points and edges are rigid bodies between adjacent joint points. The representation and storage of a graph are mostly used in the matrix as well as a 2-order tensor. Therefore, the tensor-based time series is the most natural way for expressing human action sequences since they are multi-dimensional data objects not only capturing spatial and temporal information but also preserving higher order dependencies. 

\begin{figure*}[!t]
	\centering
	\includegraphics[width=0.9\textwidth,height=5.3cm]{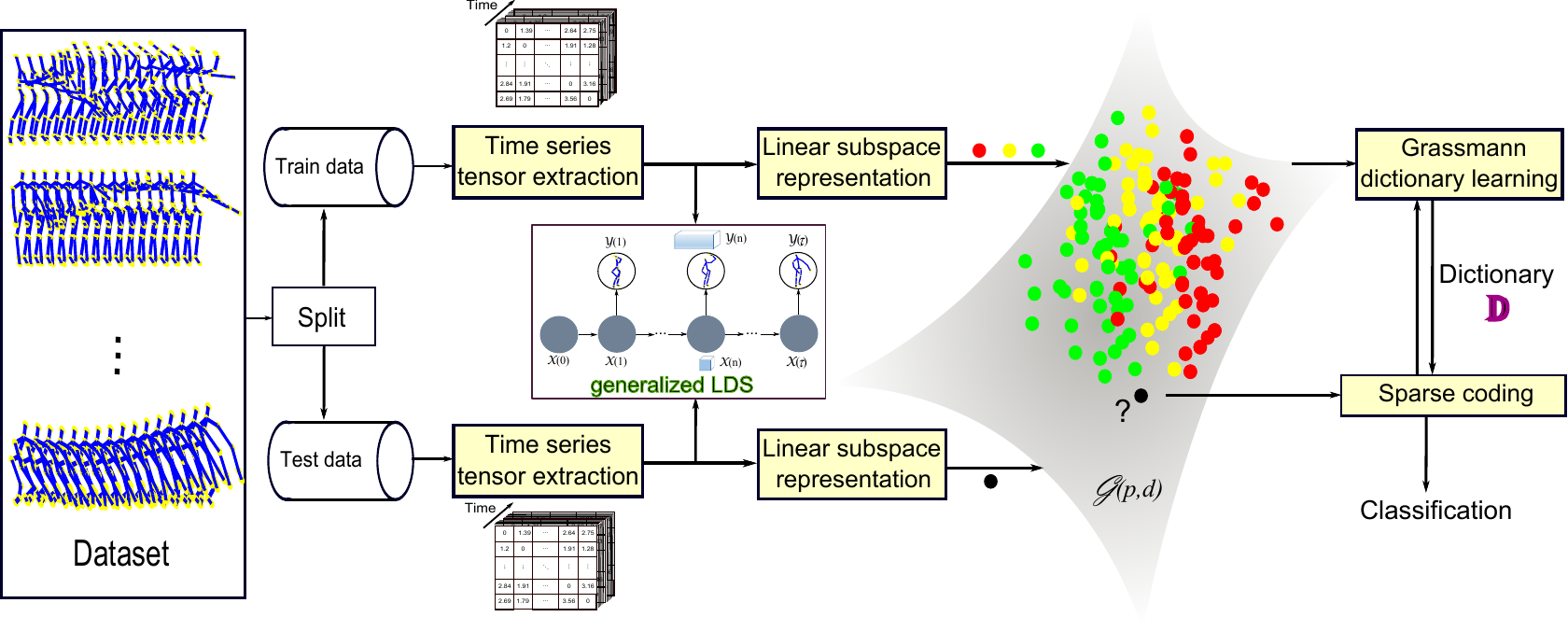}
	\caption{The general framework of the proposed approach.}  \label{fig:framework}
\end{figure*}

These ideas pose us a new way to model and compare action sequence dynamics. In order to keep the original spatio-temporal information of an action video, and improve the performance of human action recognition, this paper proposes a generalized LDS (gLDS) framework shown in Fig. \ref{fig:framework}. First, human skeletons consisted by 3D human joint points in Euclidean space are extracted from depth camera. An action video (a time series of human skeleton) is represented as a $n$-order tensor time series while each skeleton is converted to a $(n-1)$-order tensor. Based on this action representation, a tensor is decomposed into a \textit{core tensor} multiplied by a matrix along each mode. Then a subspace, spanned by columns of the observability matrix of the gLDS model, can be learned by using gLDS parameters acquired with the mode-n matricization of Tucker model. Therefore, an action can be represented as a subspace corresponds to a point on a Grassmann manifold $\mathcal{G}$. Finally, dictionary learning and sparse coding over Grassmann manifold, have been used to perform human action classification.

The contribution of the paper are the following:
(1)  We propose a novel skeleton-based tensor representation which not only keeps the original spatio-temporal information but also avoids the curse of dimensionality caused by the vectorization. 
(2) We model tensor time series utilizing gLDS model which generalizes vector-based states to tensor-based states via a multi-linear transformation. The gLDS models each tensor observation in the time series as the projection of the corresponding member of a sequence of latent tensors.
(3)  Compared to subspace methods \cite{doretto2003dynamic} the gLDS decomposes tensor-based time series to reveal the principal components which construct human skeleton. Therefore, the gLDS model achieves higher recognition accuracy for different datasets.
(4) Simulation experiments shows that proposed tensor-based representation performs better than many existing skeletal representations by evaluating it on three different
datasets. We also show that the proposed approach outperforms various state-of-the-art skeleton-based human action recognition approaches.

The remainder of this paper is organized as follows. Section \ref{section:Related Work} presents the related work. Section \ref{section:Briefs of Fundamental Concepts} briefly introduces some fundamental concepts of tensor and LDS. Section \ref{section:Proposed Method} elaborates  gLDS and describes how gLDS parameters are learned in the tensor time series. Section \ref{section:Experiments} presents our experimental results and discussion and Section \ref{section:Conclusions and Future Work} concludes this paper.

\section{Related Work}  \label{section:Related Work}

In this section we focus on the most recent methods of skeleton-based human action recognition from depth cameras. Two categories are reviewed: \textit{joint-based approaches} and \textit{dynamics descriptors-based approaches}. Joint-based approaches model the entire human skeleton by using a single feature representation, whereas dynamics descriptors-based approaches treat the skeleton sequence as 3D trajectories and model the dynamics of such time series.

$\textbf{Joint-based approaches:}$ The methods belonging to this category attempt to correlate joints locations. In order to add temporal information to this representation, Yang   \cite{yang2014effective} employed the differences of 3D joint position between the current frame and the preceding frame, and between the current frame and the initial one. Action classification was performed by using the Naive-Bayes nearest neighbor rule in a lower dimensional space constructed by using principal component analysis (PCA). Li  \cite{li2010action} employed a bag-of-3D-points graph approach to encode actions based on body silhouette points projected to the three orthogonal Cartesian planes. More complex representation is introduced in \cite{vemulapalli2014human}, where the relative 3D geometry between different rigid bodies is explicitly estimated. Their relative geometry between $\binom{N-1}{2}$ rigid body parts can be described as special Euclidean group $SE(3)$ using the rotation and translation. Therefore, the entire human skeleton in each frame can be described as a point in $SE(3)$. A sequence of skeletons is a curve in the Lie group $SE(3) \times ... \times SE(3)$.

$\textbf{Dynamics descriptors-based approaches:}$ Methods for 3D skeleton-based action recognition in this category focus on modeling the dynamics of either subsets or all the joints in the skeleton. This can be accomplished by considering linear dynamical systems (LDS) or Hidden Markov models (HMM) or mixed approaches.

Xia \cite{xia2012view} mapped 3D skeletal joints to a spherical coordinate system and computed a histogram of 3D joint locations (HOJ3D) to achieve a compact posture representation. Linear Discriminant Analysis (LDA) was used to project the HOJ3D and computed the $K$ visual words. The temporal evolutions of those visual words were modeled by a discrete Hidden Markov Model (HMM). The parameters obtained from the LDS modeling of the skeletal joint trajectories likely describe positions and velocities of the individual joints. In \cite{chaudhry2013bio}, bio-inspired shape context features were computed by considering the directions of a set of points sub-sampled over the segments in each bodypart. The temporal evolutions of these bio-inspired features were modeled using an LDS and the method learns the corresponding estimated parameters by representing the action sequence. The Local Tangent Bundle Support Vector Machine (LTBSVM) in \cite{slama2015accurate} used LDS to describe an action as a collection of time series of 3D locations of the joints. The dynamics captured by the LDS during an action sequence can be represented by the observability matrix $\textbf{O}$. The subspace spanned by columns of $\textbf{O}$ corresponded to a point on a Grassmann manifold. While class samples were presented by a Grassmann point cloud, a Control Tangent (CT) space representing the mean of each action class was learned. Each observed sequence was projected on all CTs to form a Local Tangent Bundle (LTB) representation and linear SVM was adopted to perform classification.

\section{Briefs of Fundamental Concepts} \label{section:Briefs of Fundamental Concepts}

\subsection{Tensor Notation and Operations}

In this paper, vectors (tensors of order one) are denoted by lowercase boldface symbols like $\textbf{v}$. Matrices (tensors of order two) are denoted by uppercase boldface symbols like $\textbf{A}$. High and general tensors (order three or higher) are denoted by calligraphy symbols like $\mathcal{X}$. The order of a tensor is the number of dimensions, also known as ways or modes.

Let  $\mathcal{X} \in \mathbb{R}^{I_1\times I_2\times...\times I_n}$ be an \textit{n}-order tensor with $\mathcal{X}_{i_1i_2...i_n}$ as the $(i_1,i_2,...,i_n)$th element and $I_1, I_2, ..., I_n$ are integer numbers indicating the number of elements for each dimension. The vectorization $vec(\mathcal{X})$ $\in \mathbb{R}^I$ is obtained by shaping the tensor into a vector, where $I=I_1I_2...I_{n}$ is the scalar product of the size of each dimension. In particular, the elements of $vec(\mathcal{X})$ are given by $vec(\mathcal{X})_k=\mathcal{X}_{i_1i_2...i_n}$, where $k=1+\sum\limits_{p=1}^{n}\prod\limits_{m=1}^{p-1}I_m(i_p-1)$. Unfolding $\mathcal{X}$ along the \textit{p}-mode is denoted as $\textbf{X}_{(p)}\in\mathbb{R}^{I_p\times(I_1I_2... I_{p-1}I_{p+1}... I_n)}$. The \textit{p}-mode product of a tensor $\mathcal{X}$ by a matrix $\textbf{U} \in \mathbb{R}^{J \times I_p}$ is denoted by $\mathcal{X} \times_p \textbf{U}$ and is a tensor $\mathcal{Y}\in\mathbb{R}^{(I_1\times I_2\times ... \times I_{p-1}\times J \times I_{p+1}\times... \times I_n)}$ with entries 
\begin{equation}
(\mathcal{X}\times_p \textbf{U})(i_1,...i_{p-1},j,i_{p+1},...,i_{N})=\sum_{i_p=1}^{I_p}\mathcal{X}_{i_1i_2...i_n}\cdot\textbf{U}_{ji_p}.
\end{equation} 

The tucker decomposition is a higher-order generalizations of the matrix singular value decomposition (SVD) and principal component analysis (PCA). It decomposes a tensor $\mathcal{X}$ into a core tensor $\mathcal{Z}$ multiplied (or transformed) by a matrix along each mode. Thus, we have
\begin{equation}
\mathcal{X}=\mathcal{Z}\times_1 \textbf{U}^{(1)}\times_2 \textbf{U}^{(2)}\times \dots \times_N \textbf{U}^{(N)},  
\end{equation} 
where $\mathcal{Z} \in \mathbb{R}^{J_1 \times J_2 \times ... \times J_N}$ is called the core tensor and its entries show the level of interaction between the different components. $\textbf{U}^{(n)}\in \mathbb{R}^{J_n \times I_n}$ is the factor matrices (which are usually orthogonal) and can be thought of as the principal components in each mode. A matrix representation of this decomposition can be obtained by unfolding $\mathcal{X}$ and $\mathcal{Z}$ as 
\begin{equation}
\textbf{X}_{(n)}=\textbf{U}^{(n)}\cdot\textbf{Z}_{(n)}\cdot (\textbf{U}^{(N)}\otimes  \dots \otimes \textbf{U}^{(n+1)} \otimes \textbf{U}^{(n-1)} \otimes... \otimes\textbf{U}^{(1)} )^T,
\end{equation} 
where $\otimes$ denotes the Kronecker product.

\subsection{Linear Dynamical Systems}
\label{section:Linear Dynamical Systems}

Given a time series, $\textbf{Y} = [y(1),...,y(t),...,y(\tau)]\in \mathbb{R}^{n\times \tau}$, LDS is the fundamental tools for encoding spatio-temporal data using the following Gauss-Markov process:
\begin{align} \label{equ:equation01}
\begin{cases}
y(t) &  = \textbf{C}x(t) + w(t) \hspace{1cm}  w(t)\sim N(0,\textbf{R})   \\
x(t+1) & =  \textbf{A}x(t)+v(t)  \hspace{1cm} v(t)\sim N(0,\textbf{Q}), 
\end{cases}
\end{align}
where $y(t)$ is the $n$-dimensional observed state, $x(t)$ represent of $d$-dimensional hidden state of the LDS at each time instant $t$ and $d$ represents the \textit{order} of the LDS. $\textbf{A} \in \mathbb{R}^{d \times d}$ is the \textit{transition} matrix that linearly relates the states at time instants $t$ and $t+1$ and $\textbf{C}\in \mathbb{R}^{n \times d}$ is the \textit{observation} matrix that linearly transforms the hidden state to the output $y_t$. $w \in \mathbb{R}^n$ and $v \in \mathbb{R}^d$ are noise components modeled as normal with mean equal to zero and co-variance matrix $\textbf{R} \in \mathbb{R}^{n \times n}$ and $\textbf{Q} \in \mathbb{R}^{d \times d}$ respectively. Since $\textbf{C}$ describes the spatial appearance and $\textbf{A}$ represents the dynamics, the tuple $(\textbf{A},\textbf{C})$ can be adopted as an intrinsic characteristic of the LDS model \cite{bissacco2001recognition}. Therefore, for LDS model, the goal is to learn tuple $(\textbf{A},\textbf{C})$ of LDS model given by Equ. \ref{equ:equation01}. Given the observed sequence, subspace methods in \cite{doretto2003dynamic} is widely used to learn the optimal model parameters. In this method, for seek a closed-form solution, it uses as the singular value decomposition of the observations $\textbf{Y} = \textbf{U}\varSigma \textbf{V}^T$,  as shown in \ref{fig:svd}. If $\textbf{U}_d = \textbf{U}(:,1:d)$ and $\textbf{V}_d = \textbf{V}(:,1:d)$, the estimated model parameters \textbf{A} and \textbf{C} are given by 
\begin{align}\label{equ:equation02}   
& \hat{\textbf{C}} = \textbf{U}_d, 
& \hat{\textbf{A}} = \varSigma \textbf{V}^T_d \textbf{D}_1\textbf{V}_d(\textbf{V}^T_d\textbf{D}_2\textbf{V}_d)^{-1}\varSigma^{-1} 
\end{align}
where  $\textbf{D}_1=\left[ \begin{array}{cc}   0 & 0   \\ I_{\tau -1} & 0   \\\end{array} \right]$, $\textbf{D}_2=\left[ \begin{array}{cc}   I_{\tau -1} & 0   \\ 0 & 0   \\\end{array} \right]$ and $I_{\tau -1}$ is the identity matrix of size $\tau -1$.

Since \textbf{C} denotes the spatial appearance and \textbf{A} denotes the dynamics, the extended observability matrix $\textbf{O}_{m}^{T} = [\textbf{C}^T,(\textbf{CA})^T,(\textbf{CA}^2)^T,...,(\textbf{CA}^m)^T]^T$ can be adopted as the feature descriptor for an LDS model.  The subspace spanned by columns of this finite observability matrix $\textbf{O}_{m}^{T}$ corresponds to a point on a Grassmann manifold $\mathcal{G}(p,d)$ which is a quotient space of orthogonal group $\mathcal{O}(p)$. Points on the Grassmann manifold are equivalent classes of $p \times d$ orthogonal matrices, with $d<p$, where two matrices are equivalent if their columns span the same \textit{d}-dimensional subspace.

\section{Spatio-temporal Modeling of Tensor-based Action Sequence}  \label{section:Proposed Method}

\subsection{Tensor Time Series}

\begin{figure}[!t]
	\begin{center}
		\includegraphics[height=5cm,width=0.9\linewidth]{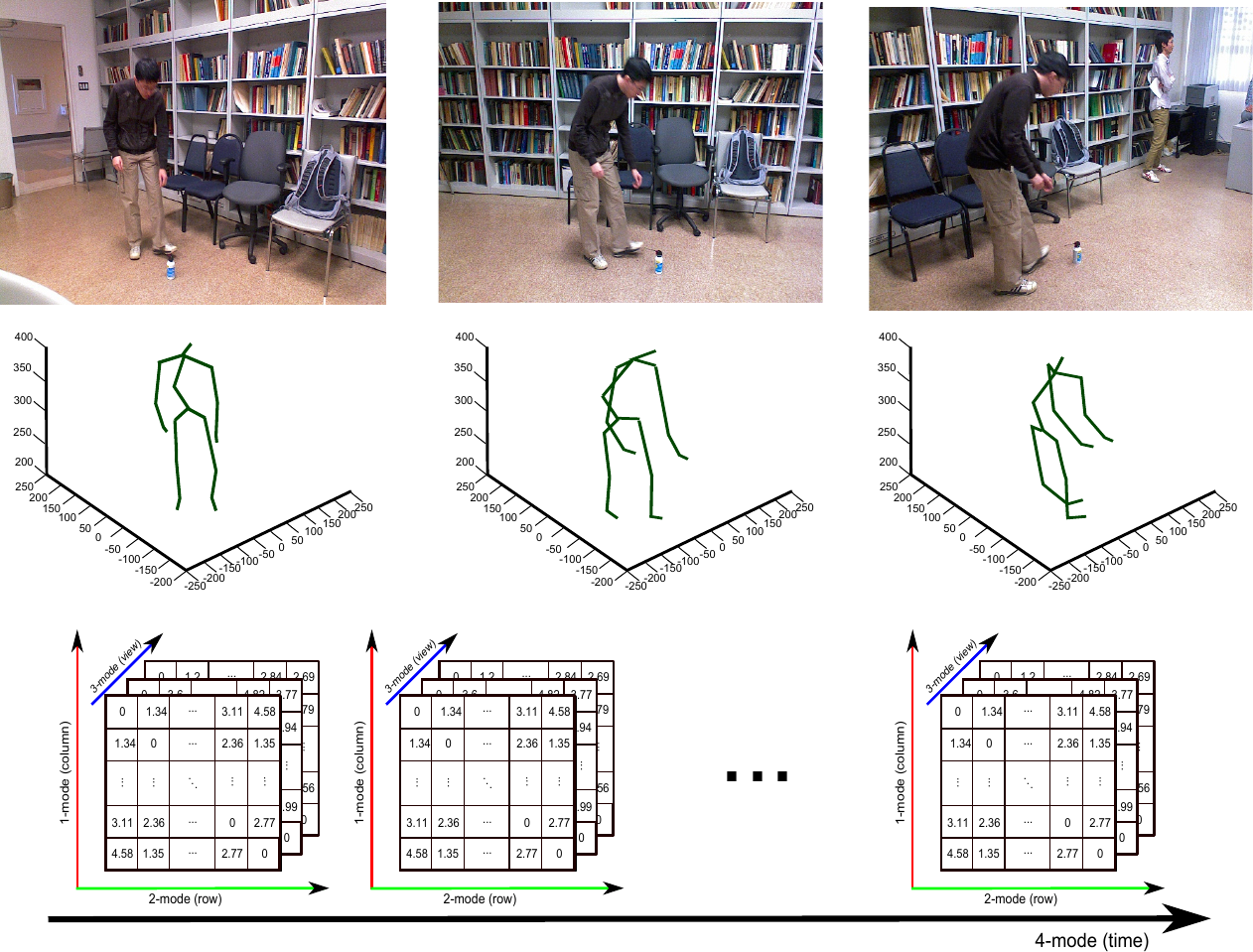} \label{fig:diff_views} 
	\end{center}
	\caption{An action sequence can be represented as a 4-order tensor while each skeleton has different views.} \label{fig:tensor_exam2}
\end{figure} 
The dynamic and continuity of movement imply that the action cannot be resumed as a simply set of skeleton because of the temporal information contained in the sequence. Instead of feature vectorization, we will consider the tensor time series, which is an ordered, finite collection of tensors that all share the same dimensionality. Such representation allows us to embody the action through searching tensor components that can better capture the variation in each mode as well as the independent of other modes.

A human skeleton has $M =N-1$ rigid bodies while it has $N$ joint points. Rigid bodies $\textbf{e}_{i,j}$ are skeleton segments between adjacent joint points $\textbf{v}_i$ and $\textbf{v}_j$, which can be described by using the direction $\textbf{v}_i-\textbf{v}_j$ of $\textbf{e}_{i,j}$ as well as $\textbf{v}_i$ and $\textbf{v}_j$. Each skeleton can be described as a 2-order tensor utilizing the following the set of rigid bodies:
\begin{align}
& \textbf{S} = [\textbf{e}_{1,2},\textbf{e}_{1,3},..., \textbf{e}_{i,j},...\textbf{e}_{N,N-1}]^{T}_{M \times 9}  
\end{align}
where $\textbf{e}_{i,j}=[\textbf{v}_i, \textbf{v}_j, \textbf{v}_i-\textbf{v}_j]$ denotes the rigid body between joint point $i$ and joint point $j$,  $\textbf{v}_i=[x_i,y_i,z_i]_{i=1:N}$ denotes the 3D position of a joint point $i$. In practice, this means that 9 parameters are needed to define the position of a 3D rigid body. A 3-order tensor time series $\mathcal{S} = [\textbf{S}_1, \textbf{S}_2,...,\textbf{S}_t,...,\textbf{S}_{\tau}]_{t=1:\tau}$ with $\tau$ frames, called 3RB, can be conveniently constructed by these orderly arrangement skeletons.

If a human skeleton has different view, an action sequence can also be indicated as a $4$-order tensor while each skeleton is represented as $3$-order tensor, as shown in Fig. \ref{fig:diff_views}. 

\subsection{Tensor-based LDS}

\begin{figure}[!t]
	\begin{center}
		\includegraphics[height=4cm,width=0.95\linewidth]{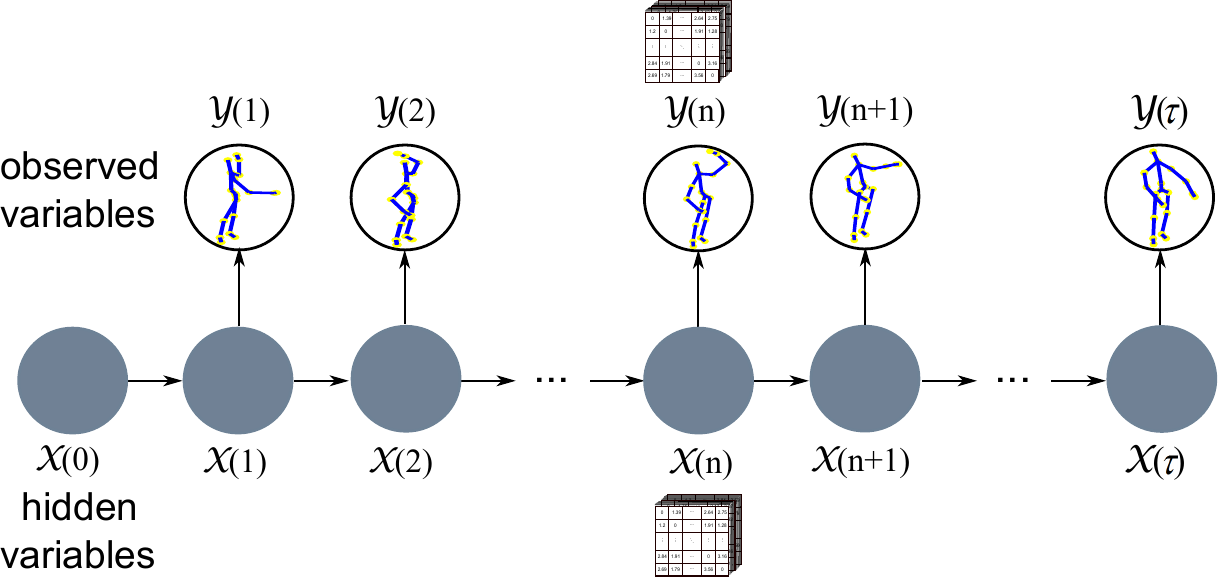} 
	\end{center}
	\caption{The gLDS model with three modes.}
	\label{fig:arma}
\end{figure}

The gLDS model presents it only for the three-way case , as shown in Fig. \ref{fig:arma}, but the generalization to $N$ ways is straightforward.
\begin{myTheo} \label{lem:Lemma01}
	Let $\mathcal{X}\in \mathbb{R}^{I_1\times I_2\times...\times I_{N}}$, $\mathcal{Y}\in \mathbb{R}^{J_1\times J_2\times...\times J_{N}}$, $\textbf{C}\in \mathbb{R}^{J\times I}$, $I = I_1I_2...I_N$, $J = J_1J_2...J_N$,  $k=1+\sum\limits_{n=1}^{N}\prod\limits_{m=1}^{n-1}I_m(i_n-1)$, $l=1+\sum\limits_{n=1}^{N}\prod\limits_{m=1}^{n-1}J_m(j_n-1)$, $1\leq i_n \leq I_n$, $1\leq n\leq N$. Then
	\begin{equation}
	\mathcal{Y}_{j_1...j_N} =(\textbf{C} \circledast \mathcal{X})_{j_1...j_N} = \sum_{l}\textbf{C}_{kl}\mathcal{X}_{i_1...i_N}.
	\end{equation} 
	\rm The product $\mathcal{Y} = \textbf{C} \circledast \mathcal{X}$ is only defined if the column number of $\textbf{C}$ matches the product of the dimension of $\mathcal{X}$. Note that this tensor product generalizes the standard matrix-vector product in the case $N=1$. We shall primarily work with tensors in their vector and matrix representations. Hence, we appeal to the following
	\begin{equation}
	vec(\textbf{C}\circledast\mathcal{X}) = \textbf{C}vec(\mathcal{X})
	\end{equation}
	Proof. $vec(\textbf{C}\circledast\mathcal{X})_k = \sum_{i_1...i_N} \textbf{C}_{kl}\mathcal{X}_{i_1...i_N} = \sum_{l}\textbf{C}_{kl}vec(\mathcal{X})_l = (\textbf{C}vec(\mathcal{X}))_k$. $\hfill\rule{3mm}{3mm}$
\end{myTheo}

In our gLDS equations, according to Lemma \ref{lem:Lemma01}, the observed states $\mathcal{Y}(t)$ and hidden states $\mathcal{X}(t)$ can be extended as tensor time series as follow:
\begin{align} \label{equ:equation04}
\begin{cases}
\mathcal{Y}(t) &  = \textbf{C} \circledast 
\mathcal{X}(t) + \mathcal{W}(t) \hspace{1cm} \mathcal{W}(t)\sim N(0,\mathcal{R})   \\
\mathcal{X}(t+1) & =  \textbf{A} \circledast \mathcal{X}(t)+\mathcal{V}(t)  \hspace{1cm} \mathcal{V}(t)\sim N(0,\mathcal{Q}), 
\end{cases}
\end{align}
where $\mathcal{Y}(t) \in \mathbb{R}^{J_1 \times J_2}$, $\mathcal{X}(t) \in \mathbb{R}^{I_1 \times I_2}$, $0 \leq t \leq \tau$, $\textbf{C} \in  \mathbb{R}^{J\times I}$, $\textbf{A} \in \mathbb{R}^{I\times I}$, $I = I_1I_2$, $J = J_1J_2$, $\mathcal{W}(t) \in \mathbb{R}^{J_1 \times J_1}$ and $\mathcal{V}(t) \in \mathbb{R}^{I_1 \times I_2}$ denote the process and measurement noise components, respectively. The goal of system identification is to estimate the parameters $\textbf{A}$ and $\textbf{C}$ from the tensor time series $\mathcal{Y}(1), ...,\mathcal{Y}(\tau)$.

\begin{figure}[!t]
	\begin{center}
		\includegraphics[height=3cm,width=0.95\linewidth]{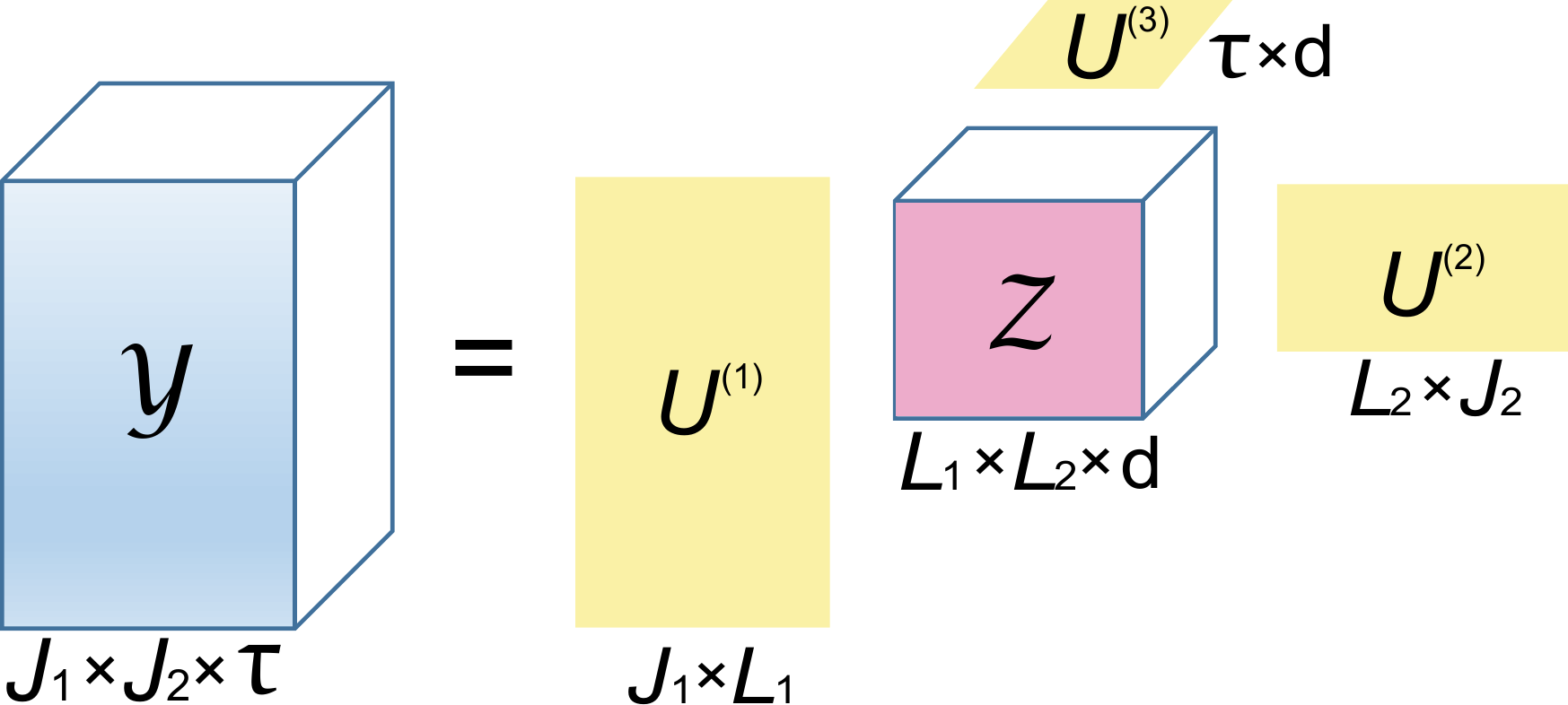} 
	\end{center}
	\caption{Tucker decomposition of a third-order tensor $\mathcal{Y}$. The column spaces of $\textbf{U}^{(1)}$, $\textbf{U}^{(2)}$ and $\textbf{U}^{(3)}$ represent the subspaces for the three modes. The core tensor $\mathcal{Z}$ is non-diagonal, accounting for possibly complex interactions among tensor components.}
	\label{fig:tucker}
\end{figure}

Let $\mathcal{Y}^{1:\tau} = [\mathcal{Y}(1), ...,\mathcal{Y}(\tau)] \in \mathbb{R}^{J_1\times J_2\times\tau}$, $\mathcal{X}^{1:\tau} = [\mathcal{X}(1), ...,\mathcal{X}(\tau)] \in \mathbb{R}^{I_1\times I_2\times\tau}$, $\mathcal{W}^{1:\tau} = [\mathcal{W}(1), ...,\mathcal{W}(\tau)] \in \mathbb{R}^{J_1\times J_2\times\tau}$  and notices that
\begin{equation}  \label{equ:equation07}
\mathcal{Y}^{1:\tau} = [\textbf{C}\circledast\mathcal{X}(1), ..., \textbf{C}\circledast\mathcal{X}(\tau)] + \mathcal{W}^{1:\tau} .
\end{equation}
Now the tucker decomposition to the tensor time series $\mathcal{Y}^{1:\tau}$ as shown in Fig. \ref{fig:tucker},
\begin{equation}  \label{equ:equation08}
\mathcal{Y}^{1:\tau} =  \mathcal{Z}\times_1\textbf{U}^{(1)}\times_2\textbf{U}^{(2)}\times_3\textbf{U}^{(3)},
\end{equation} 
where $\mathcal{Z} \in \mathbb{R}^{L_1 \times L_2 \times d}$, $\textbf{U}^{(1)} \in \mathbb{R}^{J_1 \times L_1 }$, $\textbf{U}^{(2)} \in \mathbb{R}^{J_2 \times L_2}$ and $\textbf{U}^{(3)} \in \mathbb{R}^{ \tau \times d}$. According to the specified dimensions in $(L_1, L_2, d)$, tucker decomposition computes the best rank approximation of tensor $\mathcal{Y}$, where $1 \leq L_1 \leq J_1$, $1 \leq L_2 \leq J_2$ and $1 \leq d \leq \tau$. Tucker decomposition is to reveal the latent semantic associations between human skeleton changes over time. Then, we consider the special case of mode-(3) unfolding to the Equ. (\ref{equ:equation07}) and (\ref{equ:equation08}).
\begin{align} \label{equ:equation05}
\textbf{Y}_{(3)}&  =  \textbf{U}^{(3)}\textbf{Z}_{(3)}(\textbf{U}^{(2)}\otimes\textbf{U}^{(1)})^T \notag \\
& = [vec(\textbf{C}\circledast\mathcal{X}(1)), ..., vec(\textbf{C}\circledast\mathcal{X}(\tau))]^T + \textbf{W}_{(3)}  \notag\\
& = [\textbf{C} vec(\mathcal{X}(1)), ..., \textbf{C}vec(\mathcal{X}(\tau))]^T + \textbf{W}_{(3)} \notag \\
& = \textbf{X}_{(3)}\textbf{C}^T + \textbf{W}_{(3)},
\end{align} 
where $\textbf{Y}_{(3)} \in \mathbb{R}^{\tau \times J}$, $\textbf{Z}_{(3)}  \in \mathbb{R}^{d \times L}$  and $\textbf{W}_{(3)}  \in \mathbb{R}^{\tau \times J}$ are the mode-(3) unfolding of the tensor $\mathcal{Y}^{1:\tau}$, $\mathcal{Z}$ and $\mathcal{W}^{1:\tau}$ respectively. Transpose on both sides of the Equ. \ref{equ:equation05}, we have
\begin{align} \label{equ:equation06}
\textbf{Y}_{(3)}^T & = (\textbf{U}^{(2)}\otimes\textbf{U}^{(1)})(\textbf{U}^{(3)}\textbf{Z}_{(3)})^T \notag \\
& = \textbf{C}[vec(\mathcal{X}(1)), ..., vec(\mathcal{X}(\tau))] + \textbf{W}_{(3)}^T \notag \\
& = \textbf{C}\textbf{X}_{(3)}^T+ \textbf{W}_{(3)}^T.
\end{align} 

From the definition to unfolding of tensor, we notice the equation $\textbf{Y}_{(3)}^T = [vec(\mathcal{Y}(1)), ..., vec(\mathcal{Y}(\tau))]$ and $\textbf{X}_{(3)}^T = [vec(\mathcal{X}(1)), ..., vec(\mathcal{X}(\tau))]$. Now consider the problem of finding the best estimate of $\hat{\textbf{C}} \in \mathbb{R}^{J\times I}$  in the sense of Frobenius: $\hat{\textbf{C}}$, $\hat{\textbf{X}}_{(3)}^T$ = $argmin_{\textbf{C},\textbf{X}_{(3)}^T}||\textbf{W}_{(3)}^T||$ subject to Equ. \ref{equ:equation06}. It follows immediately from the fixed rank approximation property of tucker decomposition that an estimation of the subspace mapping matrix and the underlying state sequence is given by setting
\begin{align} 
& \hat{\textbf{C}}  = \textbf{U}^{(2)}\otimes\textbf{U}^{(1)} \\
& \textbf{X}_{(3)}^T = (\textbf{U}^{(3)}\textbf{Z}_{(3)})^T. \label{equ:equation8}
\end{align}

\begin{algorithm}[H]
	\label{key}  \caption{Learning the gLDS model with n-order tensor time series}  \label{algo:al1}
	\begin{algorithmic}[1]
		\REQUIRE  n-order tensor time series $\mathcal{Y}$, dimension of subspaces $d$ and the truncation parameter of observation $m$
		\ENSURE the subspace $\textbf{S}$ correspond to $\mathcal{Y}$
		
		\STATE $\mathcal{Y} =  \mathcal{G}\times_1\textbf{U}^{(1)}\times_2\textbf{U}^{(2)}\times_3...\times_n\textbf{U}^{(n)}$; $*$ \textit{Tucker decomposition of $\mathcal{X}$}
		\STATE $\textbf{Y}_{(n)}^T = (\textbf{U}^{(n-1)}\otimes\textbf{U}^{(n-2)} \otimes... \otimes\textbf{U}^{(1)})(\textbf{U}^{(n)}\textbf{G}_{(n)})^T$
		\STATE $\hat{\textbf{C}}  \longleftarrow \textbf{U}^{(n-1)}\otimes\textbf{U}^{(n-2)} \otimes... \otimes\textbf{U}^{(1)}$
		\STATE $\textbf{X}_{(n)}^T = (\textbf{U}^{(n)}\textbf{Z}_{(n)})^T$
		\STATE $\hat{\textbf{A}}   \longleftarrow  \textbf{X}_{(n)}^T(:,2:\tau) \textbf{X}_{(n)}^T(:,1:\tau-1)^{\dagger} $
		\STATE $\textbf{O}  \longleftarrow [\hat{\textbf{C}}^T,(\hat{\textbf{C}}\hat{\textbf{A}})^T,(\hat{\textbf{C}}\hat{\textbf{A}}^2)^T,...,(\hat{\textbf{C}}\hat{\textbf{A}}^m)^T]^T$
		\STATE $ \textbf{O} = \textbf{U} \sum V^T$ $*$ \textit{compute SVD of $\textbf{O}$   }
		\STATE $\textbf{S}  \longleftarrow \textbf{U}(:,1:d)$      
	\end{algorithmic}
\end{algorithm}

Then $\hat{\textbf{A}} \in \mathbb{R}^{I\times I}$ can be determined uniquely, again in the sense of Frobenius, by solving the following linear problem: $\hat{\textbf{A}}  = argmin_{\textbf{A}}||\textbf{X}_{(3)}^T(:,2:\tau) - \textbf{A}\textbf{X}_{(3)}^T(:,1:\tau-1)||_F$, which is trivially done in closed-form using the state
estimated from Equ. \ref{equ:equation8}:
\begin{align} \label{equ:equation09}
\hat{\textbf{A}} & =\textbf{X}_{(3)}^T(:,2:\tau) \textbf{X}_{(3)}^T(:,1:\tau-1)^{\dagger} 
\end{align} 
where $\dagger$ denotes the Moore-Penrose inverse. Given the above estimates of $\hat{\textbf{C}}$ and $\hat{\textbf{A}}$, the covariance matrices $\hat{\textbf{Q}}$ and $\hat{\textbf{R}}$ can be estimated directly from residuals.

Starting from the initial state $\mathcal{X}_1$, the expected  observation sequence of gLDS is obtained as 
\begin{equation}  \label{equ:equation10}
\begin{split}
&E[\mathcal{(Y)}_1,\mathcal{(Y)}_2,\mathcal{(Y)}_3,...]=[\hat{\textbf{C}}^T,(\hat{\textbf{C}}\hat{\textbf{A}})^T,(\hat{\textbf{C}}\hat{\textbf{A}}^2)^T,...]\circledast \mathcal{X}_1 \\
&=\mathcal{O}(d,\infty)\circledast\mathcal{X}_1 \hspace{0.6cm}
s.t. \hspace{0.4cm} \hat{\textbf{C}}^T\hat{\textbf{C}}=I;  \hspace{0.2cm}  |\mu(\hat{\textbf{A}})|<1   
\end{split}
\end{equation}
where $\mu(\hat{\textbf{A}})$ denotes an arbitrary eigenvalue of $\hat{\textbf{A}}$. The transition matrix $\hat{\textbf{A}}$ needs to be stable with eigenvectors inside the unit circle. Therefore, we utilize a constraint generation method \cite{Siddiqi20075871}, which achieves a stable result efficiently by iteratively checking the stability criterion and generating new constraints. As proposed in \cite{bissacco2001recognition} the extended observability can be approximated by taking $m$-order observability matrices, which can be written as $\mathcal{O}(m) = \{\textbf{O} | \textbf{O}=[\hat{\textbf{C}}^T,(\hat{\textbf{C}}\hat{\textbf{A}})^T,...,(\hat{\textbf{C}}\hat{\textbf{A}}^m)^T]$. In this way, an action can be alternately identified as an $d$-dimensional subspace of $\mathbb{R}^{Jm \times d}$. Thus, given a database of videos, we estimate parameters of gLDS as described above for each video. Since, an action video can be alternately identified as the subspace spanned by columns of $\mathcal{O}(m)$ corresponds to a point on a Grassmann manifold. Algorithm \ref{algo:al1} provides the pseudo-code for extraction subspace using gLDS with n-order tensor time series. 

\subsection{Discussion}

\begin{figure}[!t]
	\centering
	\subfloat[]{\includegraphics[height=2cm,width=0.9\linewidth]{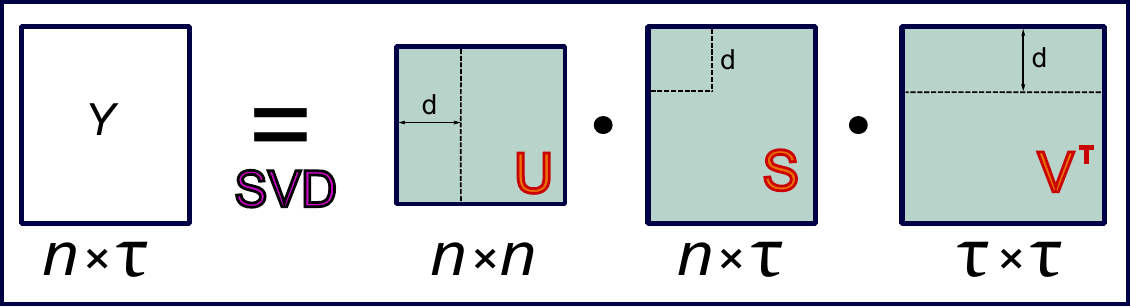}\label{fig:svd}} \\
	\subfloat[]{\includegraphics[height=2cm,width=0.9\linewidth]{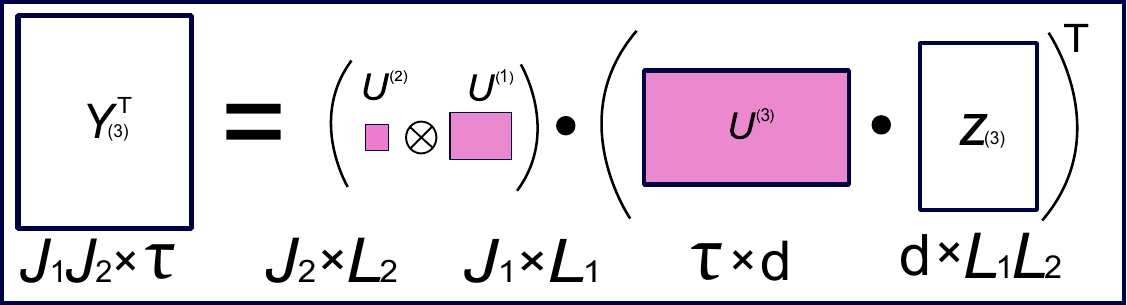} \label{fig:tucker_matrix}} 
	\caption{(a) Standard SVD of a matrix \textbf{Y} and its components \textbf{U}, \textbf{V} (unitary matrices) and \textbf{S} (diagonal matrix). The dashed lines indicate the row/column truncation. (b) The standard SVD can be replaced by tensor decomposition and unfolding. After this, $Y^T_{(3)}$ is equal with $Y$ in its dimension.} 
\end{figure}

The method introduced in \ref{section:Linear Dynamical Systems} is shown to be a valid approach to learn the parameters $(\textbf{A}, \textbf{C})$ of LDS model. The order-$d$ of the LDS model is related with the best rank approximation of $\textbf{Y}\in \mathbb{R}^{n\times \tau}$, which is determined by the truncation matrices that collect the first $d$ components of the SVD in the way indicated by the dashed lines in Fig. \ref{fig:svd}. For $\tau << n$, the order-$d$ will not be more than $\tau$. The $\textbf{C}$, denotes spatial appearance, is not associated with $\tau$. This causes a non-optimal estimation of $\hat{\textbf{C}}=\textbf{U}(:,1:d)$, especially while the variation scope for duration of time series has great fluctuation.

In this paper, we show that the standard SVD can be replaced by tensor decomposition and unfolding. This is a more natural and flexible decomposition, since it permits us to perform dimension reduction in the spatial structure $\textbf{U}^{(2)}\otimes \textbf{U}^{(1)}$ and temporal components $\textbf{U}^{(3)}$ of the video sequence. As shown in \ref{fig:tucker_matrix}, the spatial structure $\textbf{U}^{(1)}$ and $\textbf{U}^{(2)}$ are independent of $\tau$. The estimation of  $\hat{\textbf{C}}=\textbf{U}^{(2)}\otimes \textbf{U}^{(1)}$ is only associated with $L_1$ and $L_2$. Thus the ill-conditioned estimation of $\textbf{C}$ is avoided effectively.  

\section{Sparse coding on the Grassmann manifold}

In order to represent a subspace as the combination of few subspaces of a dictionary, a seemingly straightforward method \cite{goh2008clustering,xie2013nonlinear} is through embedding Grassmann manifolds into Euclidean spaces via the tangent bundle of the manifold. This method not only requires intensive computing but also makes its estimation numerically not accurate.  To avoid these limitations, another common extrinsic method \cite{Extrinsic2015Mehrtash} performs sparse coding and dictionary learning on Grassmann manifolds by embedding the manifolds into the space of symmetric matrices. Let $\mathcal{PG}(p,d)$ be the set of $d \times d$ idempotent and symmetric matrices of rank $p$. For any $\mathcal{X} \in \mathcal{G}(p,d)$, this projection mapping function $\Pi: \mathcal{G}(p,d) \rightarrow \mathcal{PG}(p,d)$ is performed by $\Pi(\mathcal{X})=\textbf{X}\textbf{X}^T=\hat{\textbf{X}}$, where $\mathcal{X}=span(\textbf{X})$ is optimized subspace spanned by the matrix $\textbf{X}$ to that of the orthonormal basis. Thus, an important metric, called the \textit{chordal metric}, will be used in a more general space to recast the problem which is hard to define tractable arithmetical calculations and distance metric on Grassmann manifold:
\begin{equation}  \label{equ:equation17}
d_{chord}(\hat{\textbf{X}},\hat{\textbf{Y}})= \parallel \Pi(\mathcal{X})-\Pi(\mathcal{Y}) \parallel_F = \parallel \hat{\textbf{X}} -\hat{\textbf{Y}}\parallel_F
\end{equation}
This metric will be used to recast the coding and consequently dictionary-learning problem in terms of chordal distance. 

Formally, given a dictionary $\mathbb{D}=\{\hat{\textbf{D}}_1,...,\hat{\textbf{D}}_j,...,\hat{\textbf{D}}_n\}$ , a query sample $\hat{\textbf{X}}$ and coefficients $\textbf{y}=[y_1,y_2,...y_N]$, $\hat{\textbf{D}}_j,\hat{\textbf{X}} \in \mathcal{PG}(p,d) $, the coding objective function with a penalty term can be written as:
\begin{equation}  \label{equ:equation16}
l(\mathcal{X},\mathbb{D}) \cong min_y \parallel \hat{\textbf{X}} - \sum_{j=1}^N y_j\hat{\textbf{D}}_j\parallel_F^2 + \lambda \parallel \textbf{y} \parallel_1.
\end{equation}
Here, The $l_1$-norm regularization is employed to the coefficients $\sum_{i=1}^N y_i =1$ for sparsity assurance and $\lambda$ is the sparsity penalty parameter.  We refer the reader to \cite{Extrinsic2015Mehrtash} for a general introduction to sparse coding and further mathematical details on their extrinsic solution for sparse coding and dictionary learning in the space of linear subspaces.

\section{Experiments} \label{section:Experiments}

In this section, we evaluate the proposed gLDS with tensor time series testing on three different datasets: MSR-Action3D \cite{li2010action}, UTKinect-Action \cite{xia2012view} and Northwestern-UCLA Multiview Action3D Dataset \footnote{http://users.eecs.northwestern.edu/$\sim$jwa368 /my$\_$data.html} .

\subsection{Alternative Representations of Tensor time Series }

To show the effectiveness of the proposed gLDS model, we compare the performance of the following four representations of tensor time series:

$\textbf{2-order joint positions  (2JP):}$ Each action sequence can be viewed as a 2-order tensor time series $3N \times \tau$, where each frame is a vector which concatenates 3D coordinates of all the joint points.

$\textbf{2-order rigid bodies (2RB):}$ Each action sequence can be viewed as a 2-order tensor time series $9(N-1) \times \tau$, where each frame is a vector which concatenates the parameters of all the rigid bodies.

$\textbf{3-order Joint positions (3JP):}$  Each action sequence can be viewed as a 3-order tensor time series $N \times 3 \times \tau$.

$\textbf{3-order Screw Motions (3SM):}$  Recently proposed in \cite{vemulapalli2014human}, screw motion between two rigid bodies is represented as point in $SE(3)$. The Lie algebra of $SE(3)$ is denoted as $\mathfrak{se}(3)$.  Mapping the point from $SE(3)$ to $\mathfrak{se}(3)$, a 6-dimensional vector representation will be acquired. Therefore, each action sequence can be viewed as a 3-order tensor time series $(N-1)(N-2) \times 6 \times \tau$.

\subsection{Evaluation Settings and Parameters}

The feature dimension depends on the number of 3D joint points (20 values for the Microsoft SDK skeleton and 15 for the PrimeSense NiTE skeleton). In the case of MSR-Action3D, UTKinect-Action datasets and Northwestern-UCLA Multiview Action3D Dataset, each skeleton has 19 rigid bodies and 20 joint points. To make the skeletal data invariant to absolute location of the human in the scene, all 3D joint coordinates is transformed from the world coordinate system to a person-centric coordinate system by placing the hip center at the origin. 

The dynamic of action is captured by using gLDS model with tensor time series. In this process, the size of core tensor $\mathcal{Z} \in \mathbb{R}^{L_1 \times L_2 \times d}$ can be significantly smaller than for the tensor time series $\mathcal{Y} \in \mathbb{R}^{J_1 \times J_2 \times \tau}$. We set $L_1 = rank(J_1)$, $L_2 = rank(J_2)$ and $d = m$ in order to approximate the value of original tensor time series, where $d$ is the subspace dimension and $m$ represents the truncation parameter of time series. Each action sequence is a point on the Grassmann manifold $\mathcal{G}_{p \times d}$ with $p = 9(N-1)m$ while skeleton is represented as 3RB. 

\begin{table}[htbp]
	\centering
	\caption{ Comparison: Recognition rate (\%) on the MSR-Action3D dataset in cross-subject setting based on AS1, AS2, and AS3.}  \label{tab:comparesubMSER}
	\begin{tabular}{p{80pt}p{20pt}p{20pt}p{20pt}p{25pt}}
		\toprule
		Method & AS1 & AS2 & AS3 &  Overall \\
		\midrule
		Bag of 3D Points\cite{li2010action}& 72.9 & 71.9 & 79.2 &  74.7     \\
		HOJ3D\cite{xia2012view}   & 88.0 & 85.5  & 63.3  &       78.9   \\ 
		Eigenjoints\cite{yang2014effective}  & 74.5 & 76.1 & 96.4 &    83.3     \\  
		LARP\cite{vemulapalli2014human}  & 95.29 & 83.87 &   98.22  &    92.46  \\       
		2JP-LDS    & 88.34 &  87.82 &  98.11 &  91.42  \\ 
		2RB-LDS    & 90.24 &  88.6 &  96.49 &  91.78  \\ 
		\hline 
		\hline 
		
		2JP-gLDS    & 95.04 &  87.17  &   98.68 &    93.63   \\ 
		2RB-gLDS    & 96.21   &    88.13   &   96.12  &    93.49   \\  
		3SM-gLDS    & 94.64 &  86.55  &   98.65 &    93.28   \\        
		3JP-gLDS    & 95.31   & 87.24  &  98.65  &  93.73  \\      
		3RB-gLDS    & \textbf{96.81}& \textbf{89.14} & \textbf{98.83}    &      \textbf{94.85}   \\         
		\bottomrule
	\end{tabular}
\end{table}

\begin{table}[htbp]
	\centering
	\caption{Recognition rate (\%) on the MSR-Action3D dataset based on experimental protocol of \cite{wang2014learning} }  \label{tab:MSR-Action3D dataset protocol}
	\begin{tabular}{|c|c|c|c|c|}
		\hline
		\bfseries  Method &   LARP \cite{vemulapalli2014human}  & LTBSVM \cite{slama2015accurate}  &  3RB-gLDS\\
		\hline
		Accuracy &      89.48   &   91.21   &  \textbf{94.96} \\
		
		\hline
	\end{tabular}
\end{table}

\subsection{MSR-Action 3D Dataset}

The MSR Action3D Dataset \cite{li2010action} is an action dataset of depth sequences captured by a depth camera. Following experimental protocol of \cite{li2010action}, the dataset was divided into subsets $AS1$, $AS2$ and $AS3$, each consisting of $8$ actions. The $AS1$ and $AS2$ group actions with similar movements, while the subset $AS3$ group complex actions with more joints engaged. We performed recognition on each subset separately using cross-subject test setting which is one half of the subjects was used as training and the other half was used as testing data. We repeated the experiment with different subspace dimension $d$ and report the mean performance. Table \ref{tab:comparesubMSER} compares the proposed approach to some methods use 3D joint positions extracted from depth videos. We can observe that the accuracy clearly outperforms the other methods. Our approach using gLDS achieves a total accuracy of 94.85$\%$ on the MSRAction3D in cross-subject experiment, significantly outperforming the other joint-based action recognition methods, including Histogram of 3D joints (HOJ3D) \cite{xia2012view}, EigenJoints \cite{yang2014effective} and Lie Algebra Relative Pairs (LARP) \cite{vemulapalli2014human}, which achieved accuracies of 78.9$\%$, 83.3$\%$ and 92.46$\%$, respectively. The average accuracy of 3RB-gLDS is 3.07$\%$ better than 2RB-gLDS. Better performance on subsets $AS1$, $AS2$ and $AS3$ indicates that the proposed gLDS is better than others in differentiating similar and complex actions.  

\begin{figure}[!t]
	\centering
	\includegraphics[height=4cm,width=0.8\linewidth]{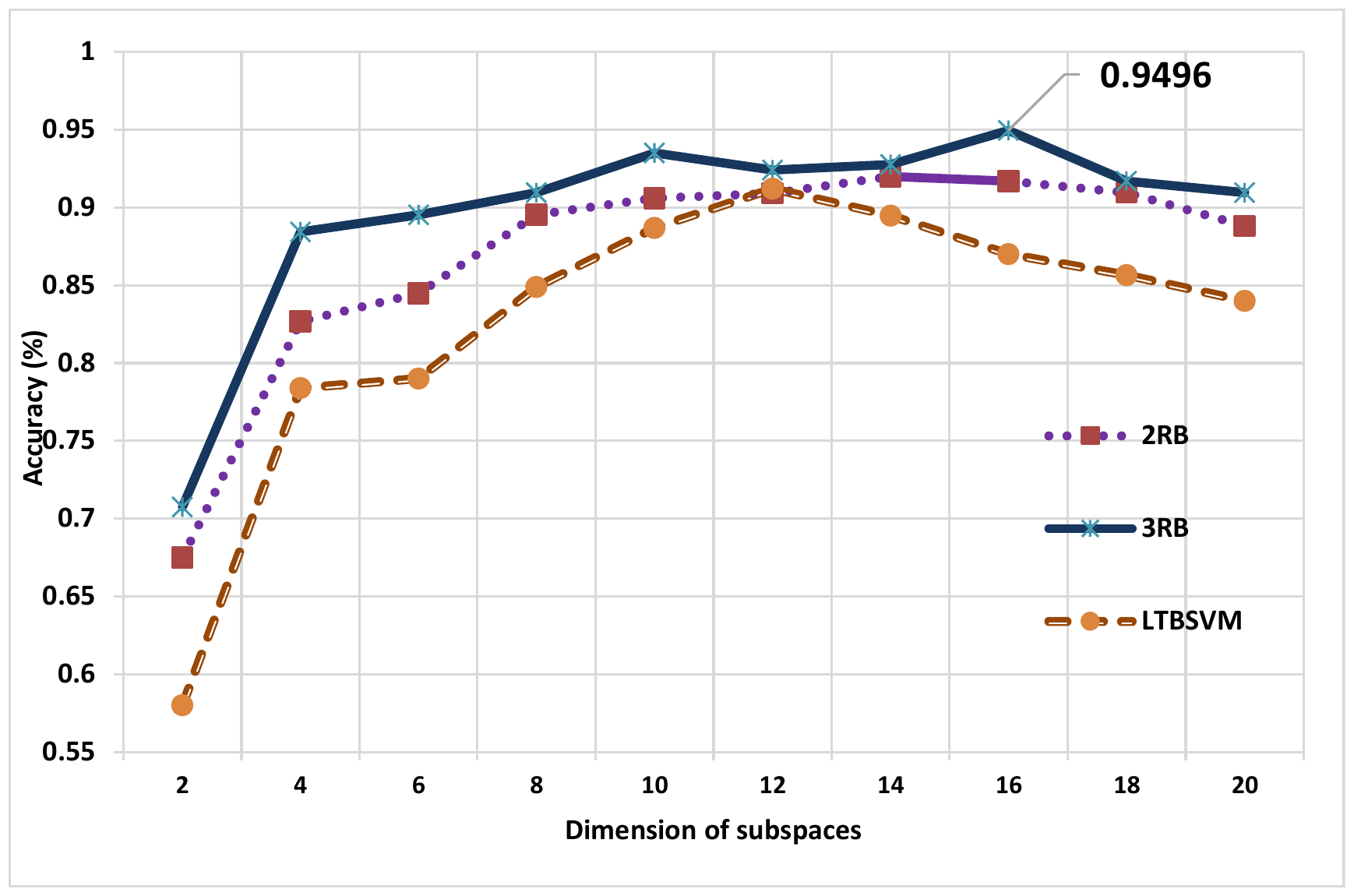}
	\caption{Recognition rate variation with learning approach and subspace dimension.}  \label{fig:dimensionSubspace}
\end{figure}

Following experimental protocol of \cite{wang2014learning}, instead of dividing the dataset into three subsets, our method achieves an total accuracy of 94.96$\%$ as shown in Tabel \ref{tab:MSR-Action3D dataset protocol}, which is applied to the entire dataset consisting of 20 actions. This experimental setting is more difficult compared to that of \cite{li2010action}. To evaluate the effect of the changing of the subspace dimensions, we conduct several tests on the MSR-action3D dataset with different dimensions of subspaces. Fig.\ref{fig:dimensionSubspace} shows the variation of recognition performances with the change of the subspace dimension. We remark that until dimension 16, the recognition rate generally increases with the increase of the size of the subspaces dimensions. This is expected, since a small dimension causes a lack of information but also a big dimension of the subspace keeps noise and brings confusion between inter-classes. We also compare in this figure, our new introduced learning algorithm 3RB to 2RB and LTBSVM \cite{slama2015accurate}.

\begin{figure}[!t]
	\centering
	\includegraphics[height=6cm,width=0.99\linewidth]{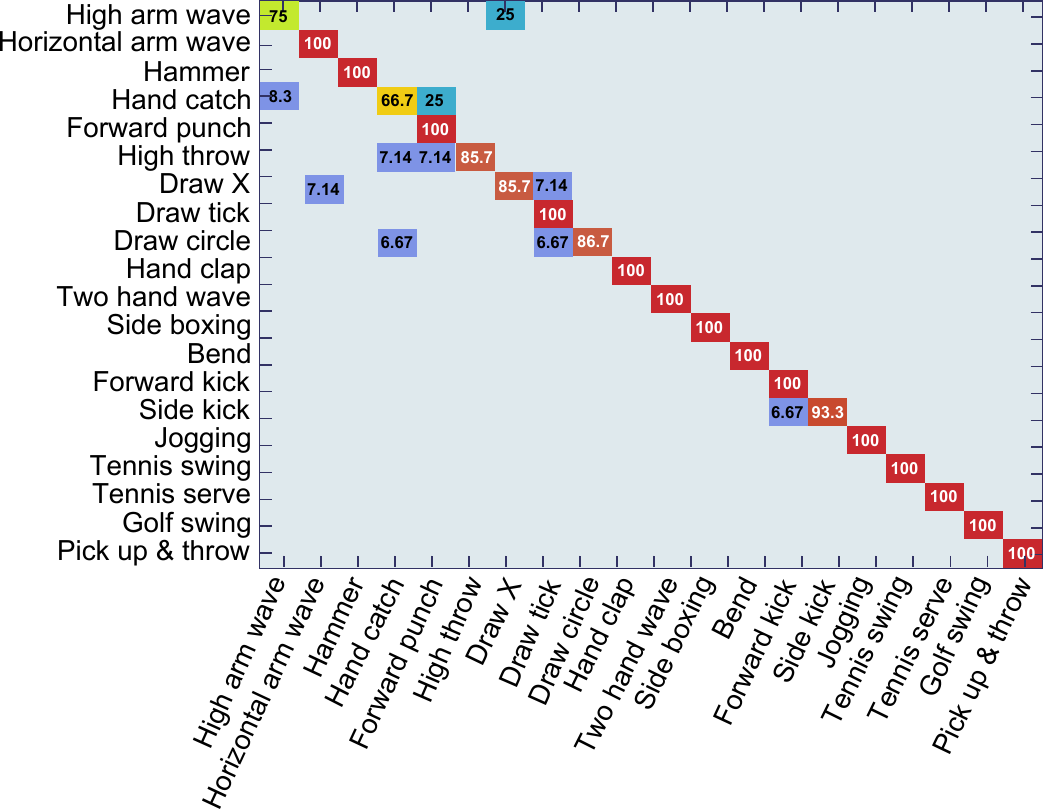}
	\caption{The confusion matrix for MSR-Action3D dataset.}  \label{fig:confusion_matrix}
\end{figure}

Figure \ref{fig:confusion_matrix} shows the confusion matrices for MSRAction3D. For most actions, about 14 classes of 20 actions, video sequences are 100$\%$
correctly classified. We can see that the
classification errors occur if two actions are too highly similar to
each other, such as \textit{high arm wave} and \textit{hand catch}.

\subsection{UT-Kinect Dataset}

Sequences of UT-Kinect dataset are taken using a stationary Kinect sensor. To allow for comparison, we followed the experiment protocol proposed by \cite{xia2012view} which is Leave One Sequence Out Cross Validation (LOOCV) on the 199 sequences. For each test, one sequence is used for testing and the other 199 sequences were used for training. Table \ref{tab:compareUTKinect} shows comparison between the recognition accuracy produced by our approach and other methods such as HOJ3D \cite{xia2012view} and LTBSVM \cite{slama2015accurate}. The accuracy of each action is more than 80$\%$ and the overall accuracy of our approach is 7.98$\%$ and 5.56$\%$ better than HOJ3D \cite{xia2012view} and LTBSVM \cite{slama2015accurate} respectively. These means that our approach is efficient and robust to change in action types thanks to the used learning approach gLDS.

\begin{table}[htbp]
	\centering
	\caption{Recognition rates  (\%) on UT-Kinect Dataset using the experiment protocol of \cite{xia2012view}}  \label{tab:compareUTKinect}
	\begin{tabular}{p{50pt}p{45pt}p{45pt}p{48pt}}
		\toprule
		Action & LITSVM\cite{slama2015accurate} & HOJ3D\cite{xia2012view} & 3RB-gLDS \\
		\midrule
		Walk      & 100 & 96.5 & 85  \\    
		Stand up  & 100 & 91.5 & 100  \\    
		Pick up   & 100 & 97.5 & 100  \\    
		Carry     & 100 & 97.5 & 95  \\    
		Wave      & 100 & 100 & 85  \\    
		Throw     & 60 & 59 & 95  \\    
		Push      & 65 & 81.5 & 100  \\    
		Sit down  & 80   & 91.5 & 100   \\    
		Pull      & 85 & 92.5 & 100  \\    
		Clap hands  & 95 & 100 & 100  \\ 
		\hline 
		\hline    
		Overall  & 88.5 & 90.92 & \textbf{96.48}   \\         
		\bottomrule
	\end{tabular}
\end{table}

\subsection{Northwestern-UCLA Multiview Action3D Dataset}

Northwestern-UCLA Multiview 3D event dataset contains RGB, depth and human skeleton data captured simultaneously by 3 Kinect cameras. Thus, each action sequence having 3 different views can be represented as a 4-order tensor time series 4RB, which the size is $(N-1) \times 9 \times 3 \times \tau $. This can help us to capture the embedded variation from different views. We follow  experiment protocol of \cite{wang2014cross} which use the samples from 9 subjects as training data, and leave out the samples from 1 subject as testing data. Tabel \ref{tab:compareMultiview} compares the recognition accuracy of our proposed gLDS and other approaches. Our method achieves higher accuracy which is about 10.8$\%$ higher than than MST-AOG \cite{wang2014cross} under the cross-subject setting. In contrast, under the cross-view setting,  the overall accuracy of our proposed method has not been greatly improved, which is only about 1.3$\%$ higher than MST-AOG \cite{wang2014cross}. This can be explained by the fact that our approach based on only 3D coordination of joint points are not enough to find view invariant features.

\begin{table}[htbp]
	\centering
	\caption{Recognition rates (\%) on Multiview-3D dataset}  \label{tab:compareMultiview}
	\begin{tabular}{p{100pt}p{50pt}p{50pt}}
		\toprule
		Method & C-Subject & C-View  \\
		\midrule
		Virtual View  \cite{li2012discriminative}& 50.7 & 47.8    \\    
		Hankelet  \cite{li2012cross} & 54.2 & 45.2    \\          
		Poselet \cite{sadanand2012action}  & 54.9 & 24.5  \\
		MST-AOG \cite{wang2014cross} & 81.6 & 73.3   \\       
		\hline 
		\hline    
		Proposed  & \textbf{92.99} & \textbf{74.6}    \\         
		\bottomrule
	\end{tabular}  
\end{table}

\section{Conclusions and Future Perspectives}  \label{section:Conclusions and Future Work}
In this paper, we have developed a novel action representation, the gLDS model, that take 3D skeleton sequence as tensor time series without unfolding the human skeletons on column vectors. It takes advantage of tucker decomposition to estimate the parameters of gLDS model as action descriptors. Our extensive experiments have demonstrated that gLDS significantly improves the accuracy and robustness for cross-subject action recognition. The major contributions include several skeleton-based tensor representation. The next step is to employ gLDS to multi-person interactions.

\section*{Acknowledgment}

This work was supported in part by the National Natural Science Foundation of China under Grant No. 61571345, the National Natural Science Foundation of China under Grant No. 9153801, and the National Natural Science Foundation of China under Grant No. 61550110247.

\section*{References}

\bibliography{TensorARMARef}

\end{document}